\documentclass[10pt,twocolumn,letterpaper]{article}

\usepackage[pagenumbers]{cvpr} 

\usepackage{lineno}
\usepackage[accsupp]{axessibility}

\renewcommand{\paragraph}[1]{\vspace{2pt}\par\noindent\textbf{#1}~}

\usepackage{multirow}

\definecolor{cvprblue}{rgb}{0.21,0.49,0.74}
\usepackage[pagebackref,breaklinks,colorlinks,allcolors=cvprblue]{hyperref}

\def\paperID{48} 
\def\confName{CVPR}
\def\confYear{2026}

\title{Neural 3D Reconstruction of Planetary Surfaces\\from Descent-Phase Wide-Angle Imagery}

\author{
\hspace{1cm}
Melonie de Almeida
\and
George Brydon
\and
Divya M.~Persaud
\hspace{1cm}
\and
\\[-12pt]
\hspace{4cm}John H.~Williamson
\and
\\[-12pt]
Paul Henderson\hspace{4cm}
\and
\\[-10pt]
University of Glasgow
}

\begin{document}
\maketitle
\begin{abstract}
Digital elevation modeling of planetary surfaces is essential for studying past and ongoing geological processes. 
Wide-angle imagery acquired during spacecraft descent promises to offer a low-cost option for high-resolution terrain reconstruction. However, accurate 3D reconstruction from such imagery is challenging due to strong radial distortion and limited parallax from vertically descending, predominantly nadir-facing cameras.
Conventional multi-view stereo exhibits limited depth range and reduced fidelity under these conditions and also lacks domain-specific priors.
We present the first study of modern neural reconstruction methods for planetary descent imaging. 
We also develop a novel approach that incorporates an explicit neural height field representation, which provides a strong prior since planetary surfaces are generally continuous, smooth, solid, and free from floating objects. This study demonstrates that neural approaches offer a strong and competitive alternative to traditional multi-view stereo (MVS) methods.
Experiments on simulated descent sequences over high-fidelity lunar and Mars terrains demonstrate that the proposed approach achieves increased spatial coverage while maintaining satisfactory estimation accuracy. 
\end{abstract}    
\section{Introduction}
\label{sec:intro}

Planetary exploration seeks to characterize the planets and moons of the solar system to understand their histories, current conditions, and the prospect of past or present extraterrestrial life. A critical step in this effort is the detailed analysis of planetary surfaces, as surface features can reveal past or present geological and hydrological activity~\cite{soderblom2007topography}. Surface reconstruction and Digital Elevation Models (DEMs) play a central role in this process~\cite{Hynek2010}, enabling scientists to construct accurate 3D models of surface elevation from remote sensing data acquired by orbiters, surface robots, or landers. The primary approach to reconstructing planetary surfaces uses stereo image pairs captured from orbit, a technique that has enabled global three‑dimensional mapping of both the Moon and Mars~\cite{Scholten2012GLD100,UCLMars3DOrbital}. However, this requires expensive cameras and orbital platforms that can prohibit its use for smaller missions, distant bodies, or new locations.

Descent imagery captured by landers can be used as a low-cost alternative to construct DEMs~\cite{soderblom2007topography,brydon2023planetary}, particularly as new exploration modes such as drones and small descent craft are being developed. Additionally, this bridges the resolution gap between the orbiter/probe and the lander's local imaging. The orbiter is far and only resolves at lower resolution, while the lander gives fine detail but only of its immediate surroundings. These findings are scientifically valuable because they give a regional context for the local science that the lander will perform, such as imaging, collecting samples, and geochemical experiments. Furthermore, the usage of descent imagery for DEM reconstruction enables using archival data from pre-landing imaging campaigns~\cite{Daubar2024,Phillips2024EuropaRecon} to reconstruct DEMs for further analysis.

\begin{figure*}[t]
    \centering
    \includegraphics[width=0.18\textwidth,height=0.18\textwidth]{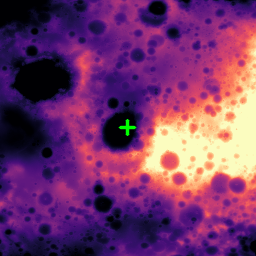}
    \hfill
    \includegraphics[width=0.18\textwidth,height=0.18\textwidth]{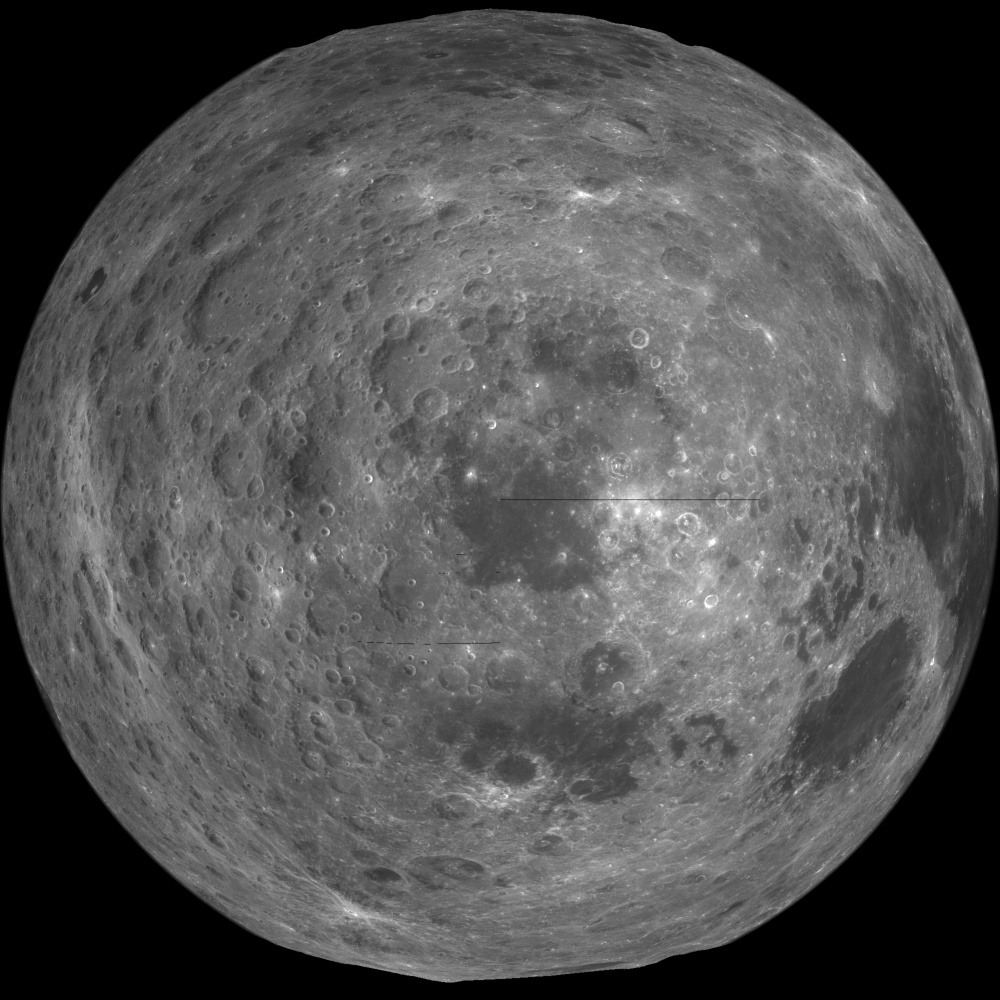}
    \hfill
    \includegraphics[width=0.18\textwidth,height=0.18\textwidth]{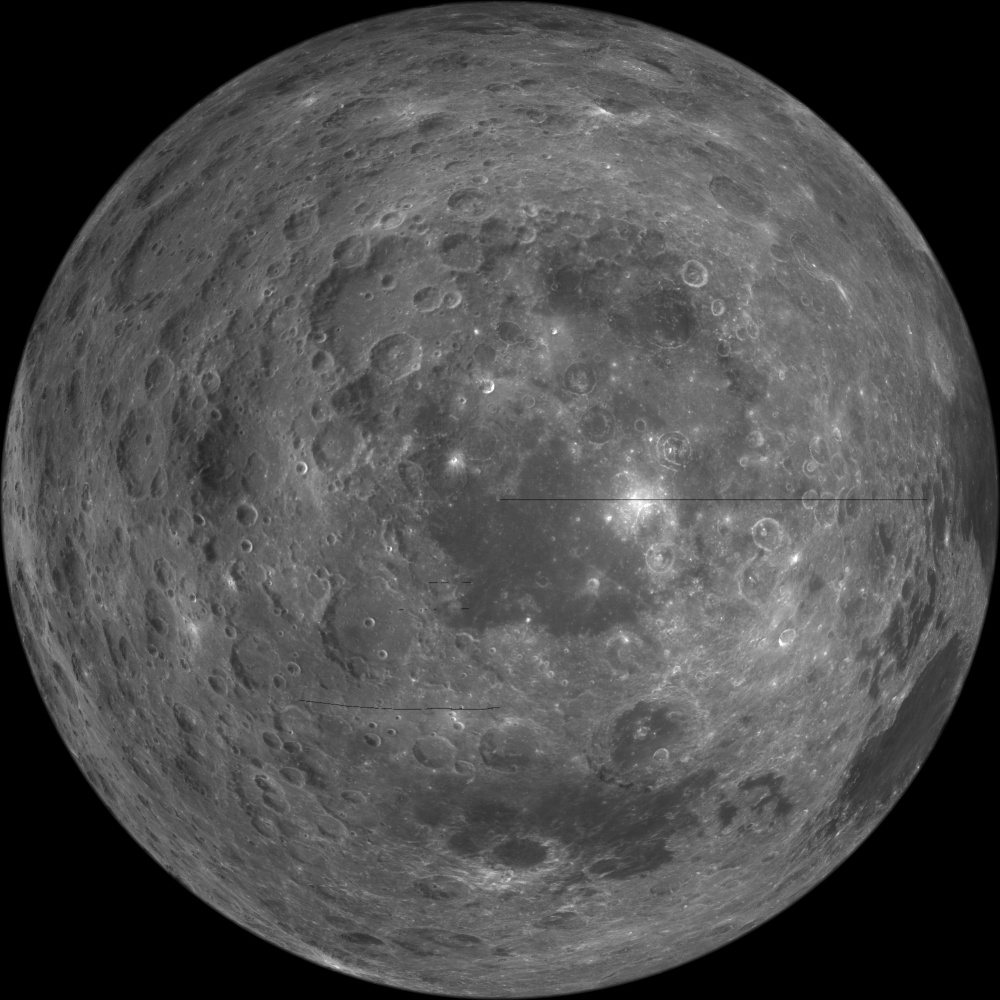}
    \hfill
    \includegraphics[width=0.18\textwidth,height=0.18\textwidth]{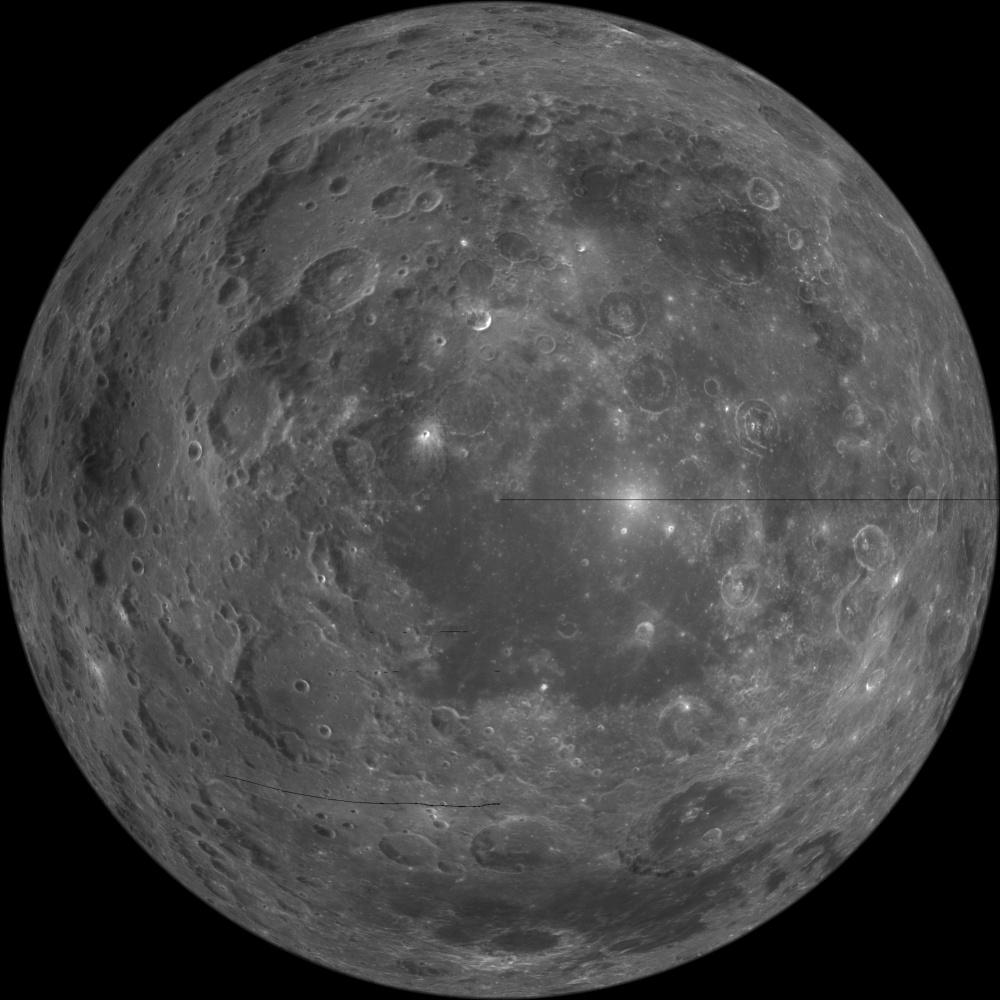}
    \hfill
    \includegraphics[width=0.18\textwidth,height=0.18\textwidth]{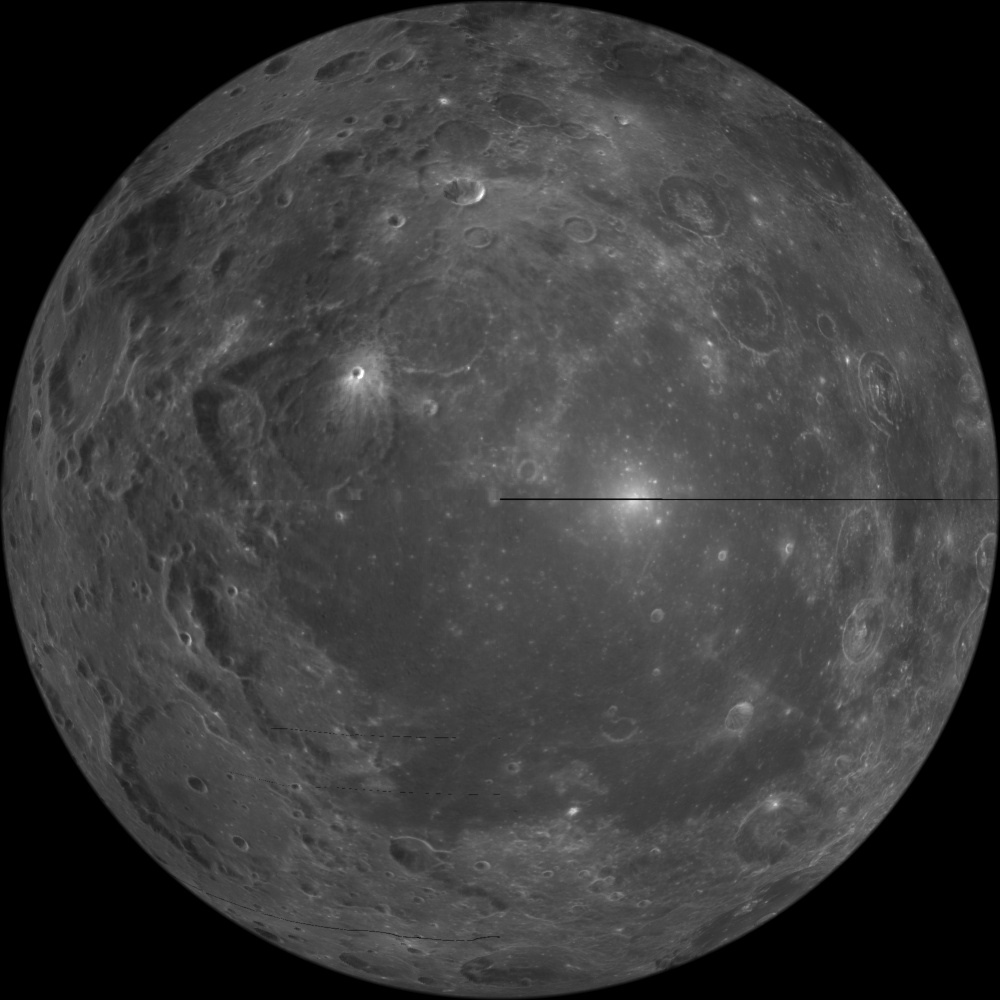}
    \caption{
    \textbf{First:} Height map and x-y camera center (green cross) for the lunar scene. \textbf{Remaining:} Example simulated descent views toward the center of the lunar scene, including the top-most and bottom-most views.
    }
    \label{fig:lunar_data}
\end{figure*} 

\begin{figure*}[t]
    \centering
    \includegraphics[width=0.18\textwidth,height=0.18\textwidth]{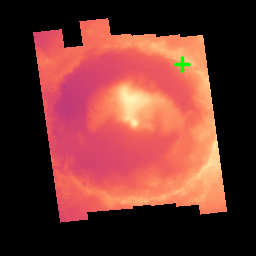}
    \hfill
    \includegraphics[width=0.18\textwidth,height=0.18\textwidth]{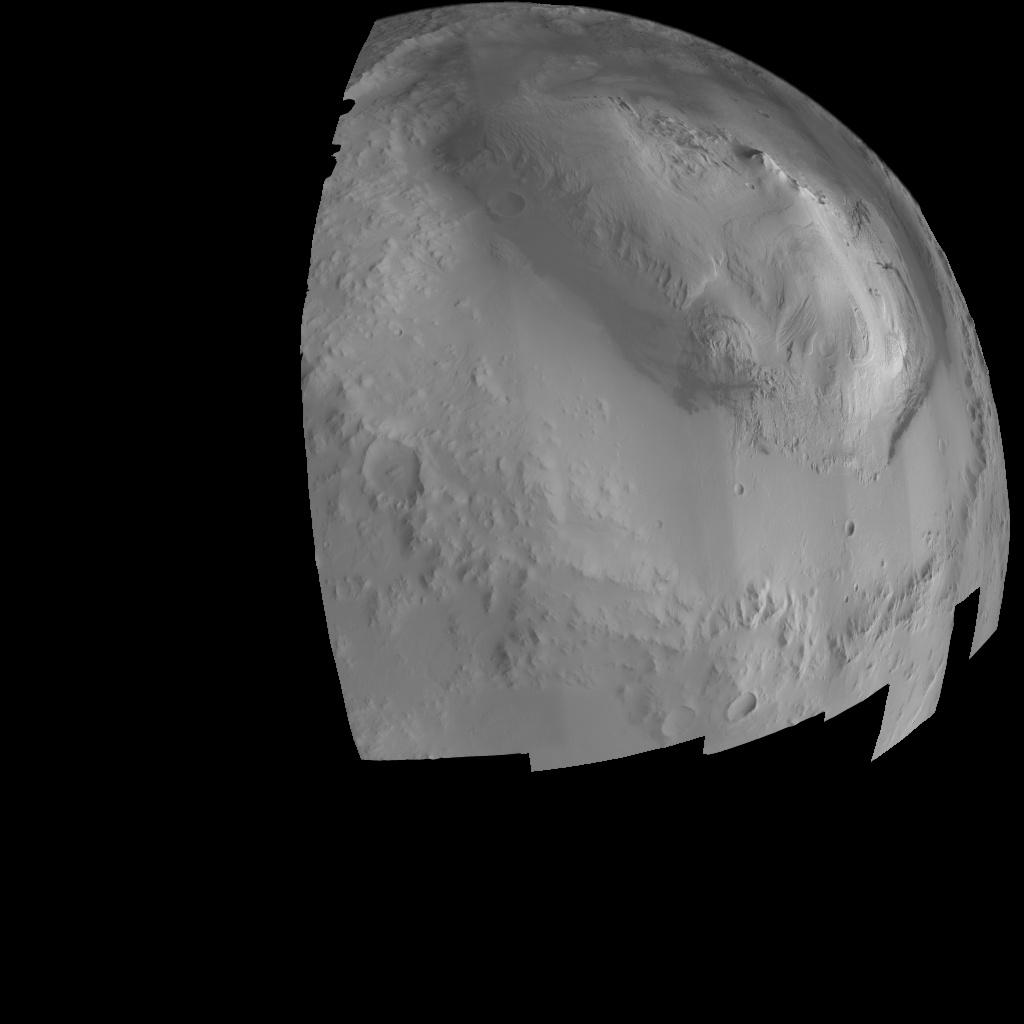}
    \hfill
    \includegraphics[width=0.18\textwidth,height=0.18\textwidth]{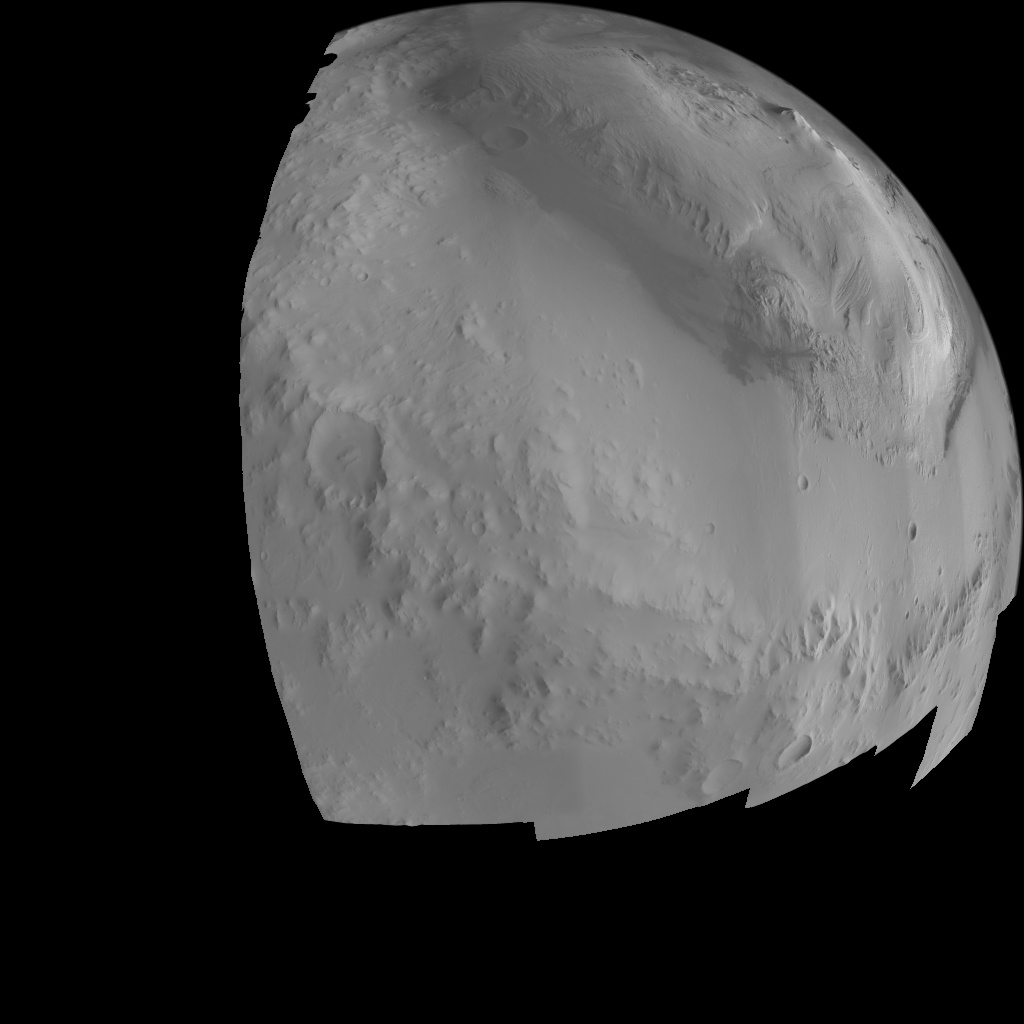}
    \hfill
    \includegraphics[width=0.18\textwidth,height=0.18\textwidth]{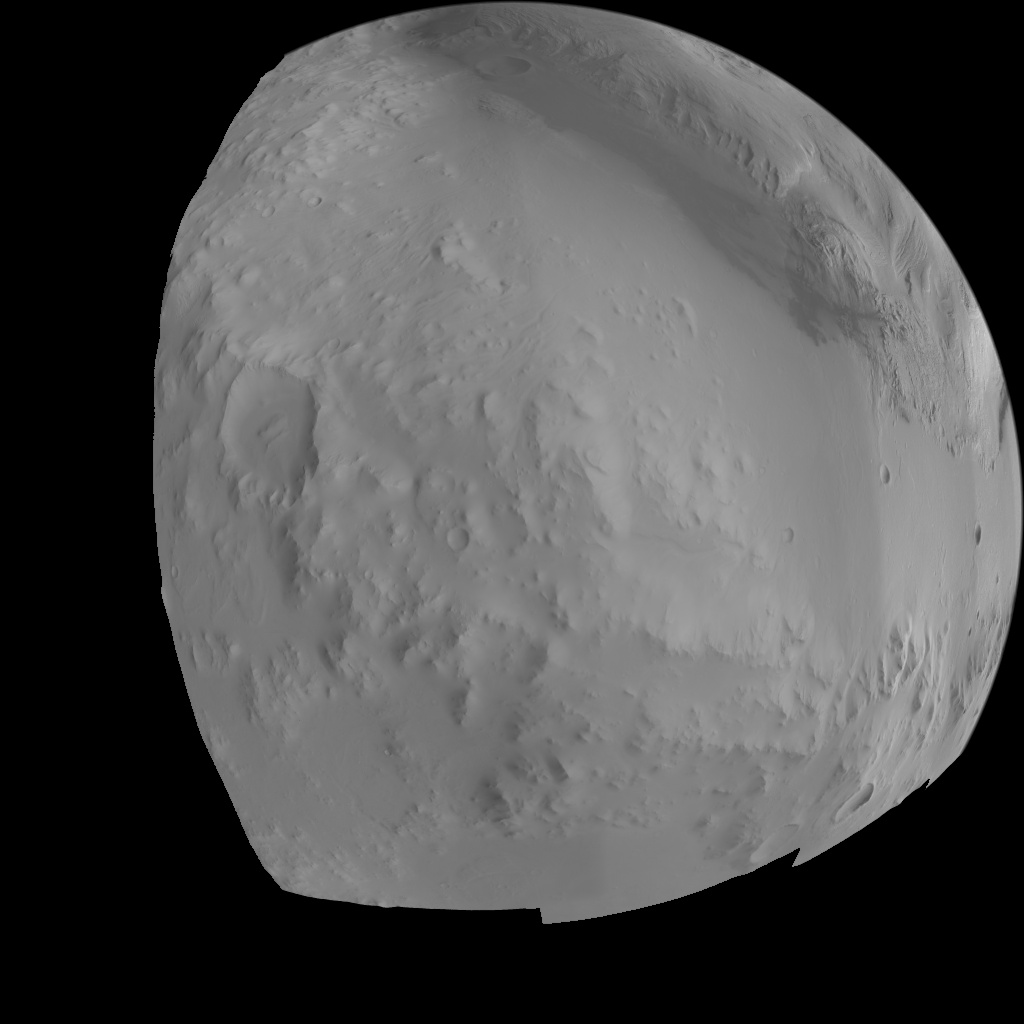}
    \hfill
    \includegraphics[width=0.18\textwidth,height=0.18\textwidth]{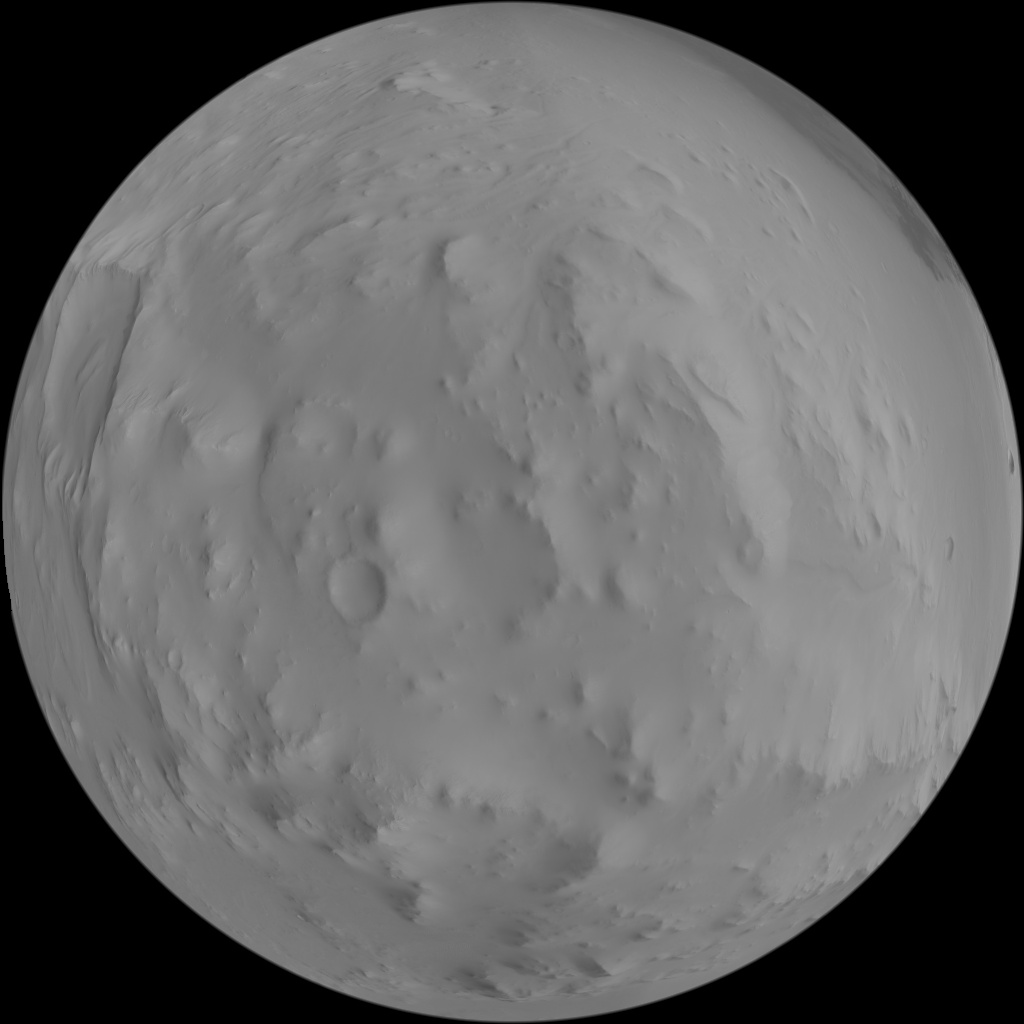}
    \caption{
    \textbf{First:}Height map and x-y camera center (green cross) for the Mars scene. \textbf{Remaining:} Example simulated descent views toward the top-right corner, including the top-most and bottom-most views.
    }
    \label{fig:mars_data}
\end{figure*} 

A compelling example comes from data of Titan by the Cassini-Huygens mission~\cite{soderblom2007topography}. River valleys have been detected at nearly all latitudes on Titan, revealing a surprisingly Earth-like landscape shaped not by water, but by liquid methane. To better understand the geomorphological processes responsible for carving these landscapes, researchers have used MicMac~\cite{rupnik2017micmac} structure-from-motion (SfM)~\cite{ullman1979interpretation} and multi-view stereo (MVS)~\cite{seitz2006comparison} to reconstruct DEMs from wide‑angle descent imagery.

Classical stereo-based pipelines often yield sparse or noisy reconstructions and cannot capture view-dependent effects such as reflections (indeed, such effects typically reduce their accuracy).
They are also sensitive to parameters and image coverage.
While modern, integrated implementations such as COLMAP~\cite{schonberger2016structure} and Agisoft Metashape~\cite{AgisoftMetashape} have made these pipelines easier to use on high-quality imagery with large overlaps.
Extreme imaging conditions, such as wide-angle images of vertical descents, can result in incomplete 3D reconstructions.

Modern NeRF-based~\cite{Mildenhall2020-hc} approaches instead learn a continuous volumetric representation, producing denser, smoother, and more accurate reconstructions that also capture viewpoint-dependent effects. NeRF-based methods replace traditional MVS by reconstructing detailed scene geometry and synthesizing novel views, taking posed images as input. This is achieved by optimizing a neural radiance field that maps 3D coordinates and viewing directions to colour and density. Several recent extensions to this approach have been proposed to support unbounded scenes~\cite{barron2021mip,barron2022mip}.

However, scenarios involving wide-angle (fisheye) images and cameras descending vertically toward flat surfaces also present challenges for NeRF-based methods. Wide-angle lenses introduce strong radial distortion that must be properly modeled for accurate ray generation. Predominantly downward-facing cameras imaging a largely planar surface also result in limited viewpoint diversity, reduced parallax, and depth ambiguity~\cite{Mildenhall2020-hc}. This makes it difficult for NeRF-based methods to infer depth and fine geometric details reliably.

In this study, we apply NeRF-based methods to construct digital elevation models (DEMs) of planetary surfaces using wide-angle descent imagery for the first time. We choose the Nerfacto model~\cite{nerfstudio2023}, a recent NeRF-based framework, as our baseline due to its computational efficiency and ability to handle unbounded scenes. However, planetary surfaces are generally smooth and continuous, devoid of floating debris, and illuminated primarily by distant sunlight. In contrast, Nerfacto is a general‑purpose NeRF formulation that does not explicitly enforce the smooth, continuous geometry and illumination conditions typical of planetary surfaces, which can lead to floating artifacts and geometric inconsistencies. To avoid these limitations and leverage domain-specific geometric and photometric priors, we propose a novel NeRF-based framework. 
It integrates an intermediate $\mathrm{HeightField}$ representation that maps 2D coordinates to elevation, enabling high-resolution planetary surface height-map generation directly from fisheye imagery. Instead of using a direction-encoded colour MLP, we adopt the physically-based Hapke reflectance model~\cite{Sato2014Hapke} to achieve photometrically-consistent and realistic appearance under extraterrestrial lighting conditions. In addition, we incorporate regularization losses to mitigate geometric ambiguity and stabilize training under limited viewpoint diversity.

Our key contributions are:
\begin{itemize}
    \item Quantitatively evaluating NeRF-based models for Digital Elevation Modeling of planetary surfaces using wide-angle descent imagery for the first time, and showing they achieve higher coverage than Metashape, while preserving satisfactory accuracy.
    \item A novel reconstruction framework that couples NeRF and heightmap representations to achieve more physically-plausible reconstructions than the Nerfacto baseline
    \item An angle-aware distortion loss to ensure smooth surface reconstruction while being flexible about the edges 
    \item Analyzing the effectiveness of incorporating a physically-based shading model specific to planetary surfaces, instead of implicitly learning direction-dependent colour.
\end{itemize}

We perform a comprehensive evaluation with simulated Fisheye imagery on existing high-resolution 3D reconstructions of lunar~\cite{Scholten2012GLD100} and Mars~\cite{UCLMars3DOrbital} surfaces. Our proposed approach achieves a balanced trade-off between spatial coverage and elevation accuracy in the generated DEMs, outperforming both Metashape~\cite{AgisoftMetashape} and Nerfacto~\cite{nerfstudio2023} in coverage at 0.1 relative error across both datasets. Moreover, it demonstrates a clear improvement over Nerfacto~\cite{nerfstudio2023} for this task, achieving lower relative and absolute elevation differences across both datasets.

\begin{figure}[t]
    \centering
    \includegraphics[width=0.5\textwidth]{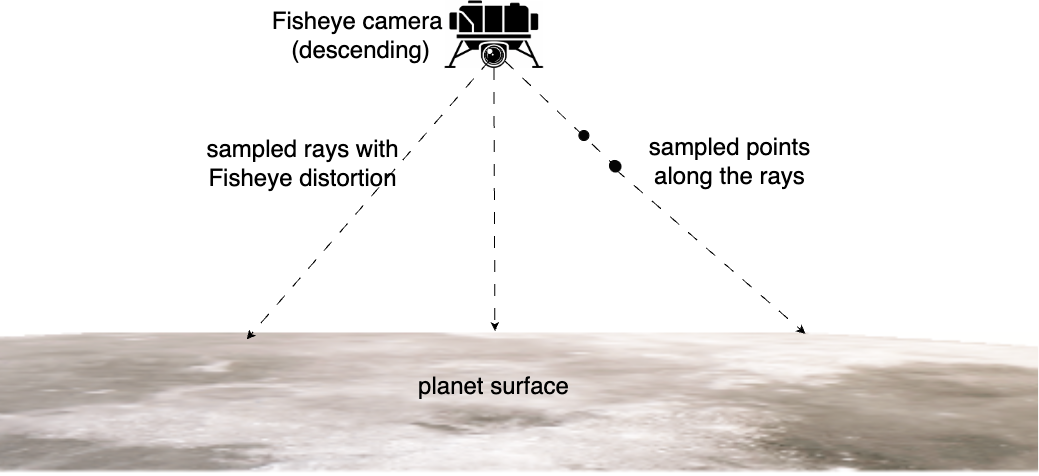}
    \vspace{-8pt}
    \caption{
    \textbf{Fisheye Ray Casting:} maps each pixel through a wide-angle lens to a 3D ray using the lens’s radial distortion model.}
    \label{fig:fisheye_rays}
\end{figure}

\section{Related Work}

\subsection{Classical 3D Reconstruction}

3D reconstruction methods can be divided into active and passive techniques~\cite{isgro2005open, zhou2024comprehensive}. Active methods emit controlled energy, such as light or laser, and analyze its interaction with surfaces; examples include structured light~\cite{rocchini2001low}, LiDAR~\cite{schwarz2010mapping}, laser scanning~\cite{kraus1998determination,gobel2007imaging}, CT scanning~\cite{flisch1999industrial}, and photometric stereo~\cite{woodham1980photometric}. Although often precise and robust, they require specialized hardware and controlled conditions.
Within passive reconstruction, traditional pipelines typically combine Structure-from-Motion (SfM)~\cite{Dellaert2003,sturm2005multi,ullman1979interpretation,sturm2006calibration,ramalingam2006generic,schonberger2016structure} and Multi-View Stereo (MVS)~\cite{786933,seitz2006comparison}. Structure-from-Motion (SfM) estimates camera poses and intrinsic parameters while simultaneously reconstructing a sparse 3D point cloud from multiple overlapping images by detecting and matching visual features across views and enforcing geometric constraints such as epipolar consistency~\cite{Dellaert2003,sturm2005multi,ullman1979interpretation,sturm2006calibration,ramalingam2006generic}. The result is a globally optimized sparse representation of the scene and calibrated camera parameters. Building on this, Multi-View Stereo (MVS) performs dense reconstruction by estimating per-pixel depth maps or dense point clouds using photometric consistency across multiple calibrated images~\cite{786933,seitz2006comparison}. MVS leverages the camera parameters obtained from SfM to recover detailed surface geometry, producing a dense 3D model suitable for surface reconstruction and DTM generation. A practical implementation of this pipeline is Agisoft Metashape, which integrates image alignment, bundle adjustment, dense reconstruction, and mesh generation into an end-to-end framework.

\subsection{3D Reconstruction with Machine Learning}

Recent methods increasingly use deep learning to improve robustness and completeness. These approaches can be broadly divided into explicit representations, such as point clouds, voxels, and meshes, and implicit representations, particularly neural implicit representations~\cite{zhou2024comprehensive}. 
Among the latter, methods based on Neural Radiance Fields (NeRF)~\cite{Mildenhall2020-hc} have demonstrated high-fidelity novel view synthesis and accurate geometry reconstruction by representing scenes as continuous volumetric radiance fields, which are learned directly from multi-view images and optimized to model both colour and density at every point in 3D space. Mip-NeRF~\cite{barron2021mip} reduces aliasing by modeling conical frustums, while Mip-NeRF 360~\cite{barron2022mip} extends this to 360° unbounded scenes using a non-linear scene parameterization and a distortion-based regularizer that concentrates density along rays, reducing floaters for more realistic reconstructions. However, these methods still face challenges in reconstructing thin surfaces.
NeRF-based methods have been applied on Earth for surface elevation modeling using images captured while moving across the surface, rather than vertically descending towards the surface~\cite{dai2024neural,abate2024application}. Alternatively, NeUS~\cite{wang2021neus} learns an implicit surface representation, instead of a density field, from multiview 2D images. However, this method cannot handle fisheye images. 
More recently, explicit point-based neural representations such as 3D Gaussian Splatting~\cite{Kerbl2023-tu,zwicker2001ewa} introduce anisotropic Gaussian primitives for scene representation, enabling real-time rendering while maintaining competitive reconstruction accuracy. However, NeRF’s volumetric representation allows it to better capture thin structures because the density field is continuous~\cite{Mildenhall2020-hc}. 
\subsection{Previous Work in Planetary Elevation Modeling using Fisheye Descent Images}

Although 3D reconstruction has recently seen significant advances with the advent of NeRF-based methods, space science continues to rely primarily on classical methods to reconstruct DEMs. Commercial software such as SOCET SET has been used to generate DEMs from descent images~\cite{soderblom2007topography} taken by the Huygens probe on Titan, enabling photogrammetric adjustment of overlapping image pairs to recover camera orientations and probe positions. More recently, IPGP-DMT leverages the open-source photogrammetry software MicMac~\cite{rupnik2017micmac} with Shape-from-Motion ~\cite{TomasiKanade1993} to produce higher-quality DEMs. MicMac’s reconstruction quality is highly sensitive to parameter settings and image coverage, often yielding noisy or incomplete DEMs. Furthermore, surface points were generated by triangulating stereo matches between two images of different altitudes~\cite{brydon2023planetary} and evaluated by measuring errors at varying radii from the image centre. This approach is limited by its reliance on rectified images and inherently sparse coverage, often producing holes or artifacts in the reconstructed surface—issues that are not captured by the evaluation metrics.
Recently, Metashape~\cite{AgisoftMetashape}, which automates the entire SfM–MVS workflow, supporting end-to-end 3D reconstruction from image alignment through textured mesh generation, is widely used in planetary science. However, Metashape relies on high-quality, well-overlapping images, which can lead to insufficient coverage in DEMs. Although NerF-based approaches have been applied to surface reconstruction of the lunar surface using stereo pairs~\cite{hansen2024analyzing}, fisheye descent image-based surface reconstruction is unexplored.

\begin{figure}[t]
    \centering
    \includegraphics[width=0.5\textwidth]{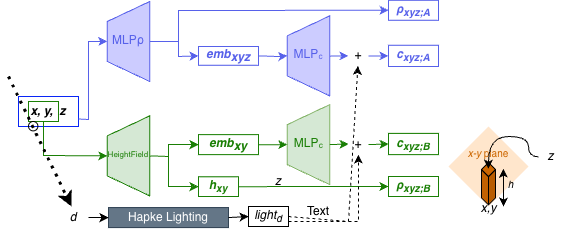}
    \vspace{-8pt}
    \caption{
    {Our method uses two scene representations that each define the density and colour of sampled points \(\mathbf{(x,y,z)}\) along the ray. In the \textbf{NeRF representation (purple)}, an \(\mathrm{MLP_{\rho}}\) maps \(\mathbf{(x,y,z)}\) into density \(\mathbf{\rho_{xyz;A}}\) and density embedding \(\mathbf{emb_{xyz}}\). \(\mathrm{MLP_{c}}\) maps  \(\mathbf{emb_{xyz}}\) into colour, which is then multiplied by ray direction dependent lighting (\(\mathbf{light_d}\)) to get final colour: \(\mathbf{c_{xyz;A}}\). In the \textbf{heightmap representation (green)}, given our world frame is defined so the surface is perpendicular to the z-axis, the \(\mathbf{(x,y)}\) is passed onto \(\mathrm{HeightField}\) to predict heights \(\mathbf{h_{xy}}\) and the density embedding of the $\mathbf{(x,y)}$ column: \(\mathbf{emb_{xy}}\). Then, $\mathbf{\rho_{xy}} = \mathbf{\rho_{xy}^{\text{const}}}$ below the height, smoothly decaying to zero via a sigmoid above it. \(\mathrm{MLP_{c}}\) maps \(\mathbf{emb_{xy}}\) into colour of the column along $z$; which is then multiplied by \(\mathbf{light_d}\) to get \(\mathbf{c_{xyz;B}}\) for each $z$ along $\mathbf{(x,y)}$ column.}}
    \label{fig:pipeline}
\end{figure} 
\section{Methodology}

In this study, we evaluate the effectiveness of NeRF-based approaches to reconstruct DEMs from wide-angle descent imagery.
We propose an extended NeRF-based approach specifically for this task to reconstruct smooth planetary surfaces while reducing the limitations of Fisheye distortion and less viewpoint diversity.

Our method reconstructs a high-resolution heightmap of the planetary surface from a fisheye image sequence, $\mathbf{I_n}$; $\mathbf{n = 1,\dots,N_\text{img}}$, acquired during the descent of the spacecraft for camera extrinsics: $[\mathbf{R}_n \mid \mathbf{t}_n]$ and intrinsics including distortion parameters: $\{\mathbf{K}_n, \boldsymbol{\xi}_n\}$. A binary mask $\mathbf{m_n}$ is provided as additional input to exclude the camera housing and any visible spacecraft structures.

\subsection{Background: Neural Radiance Fields (NeRF)}
\label{subsec:method-nerf}

NeRF~\cite{Mildenhall2020-hc,nerfstudio2023} represents a scene as a continuous volumetric field using an MLP. To render a pixel, a ray \(\mathbf{r}(t) = \mathbf{o} + t\mathbf{d}\) is cast from the camera center \(\mathbf{o}\) through the pixel along direction \(\mathbf{d}\). Distances \(\mathbf{t}\) (sorted between near plane \(t_{near}\) and far plane \(t_{far}\)) are sampled along the ray. For each point \(\mathbf{x} = \mathbf{r}(t_k)\), a  positional encoding is applied. The encoded points are fed into an MLP (which also takes view direction, omitted for brevity) to predict density \(\tau_k\) and colour \(\mathbf{c}_k\):
\begin{equation}
[\mathbf{\tau_k}, \mathbf{c}_k] = \mathrm{MLP}(\gamma(\mathbf{r}(\mathbf{t_k})); \Theta).
\end{equation}
The pixel colour is computed via numerical quadrature of the volume rendering integral:
\begin{equation}
\mathbf{C}(\mathbf{r}; \Theta, \mathbf{t}) = \sum_k T_k (1 - \exp(-\mathbf{\tau_k} \Delta_k)) \mathbf{c}_k,
\quad
\end{equation}
\begin{equation}
T_k = \exp\left(-\sum_{k' < k} \mathbf{\tau_{k'}} \Delta_{k'}\right),
\end{equation}
with \(\Delta_k = \mathbf{t_{k+1}} - \mathbf{t_k}\). Training minimizes the error between the rendered and ground-truth pixel colours. 

Nerfacto extends the standard NeRF formulation with several key modifications~\cite{nerfstudio2023}. To enable reconstruction of unbounded scenes, it employs a scene contraction function that transforms Euclidean space into a bounded domain before mapping each point ($\mathbf{x}$) into density($\tau_k$) as shown below~\cite{barron2022mip,nerfstudio2023}:
\begin{equation}
\text{contract}(\mathbf{x}) = \begin{cases}
\mathbf{x} & \|\mathbf{x}\| \leq 1 \\
\left(2 - \frac{1}{\|\mathbf{x}\|}\right)\frac{\mathbf{x}}{\|\mathbf{x}\|} & \|\mathbf{x}\| > 1
\end{cases}
\end{equation}
Furthermore, a distortion loss regularizes the ray weight distribution to be compact, suppressing floaters and discouraging background collapse~\cite{barron2022mip,nerfstudio2023}:
\begin{equation}
\mathcal{L}_{\text{dist}} = \iint w(\mathbf{u})w(\mathbf{v})|\mathbf{u-v}|\,\mathbf{du}\,\mathbf{dv};
\end{equation}
where each integral should be interpreted as a path integral along a ray.

Moreover, it uses a multi-resolution hash encoding for positional encoding, enabling fast training and high-fidelity 3D reconstruction with minimal parameter count~\cite{muller2022instant}.

\subsection{Our Heightmap Reconstruction Method}

\begin{figure*}[t]
    \centering
    \includegraphics[width=\textwidth]{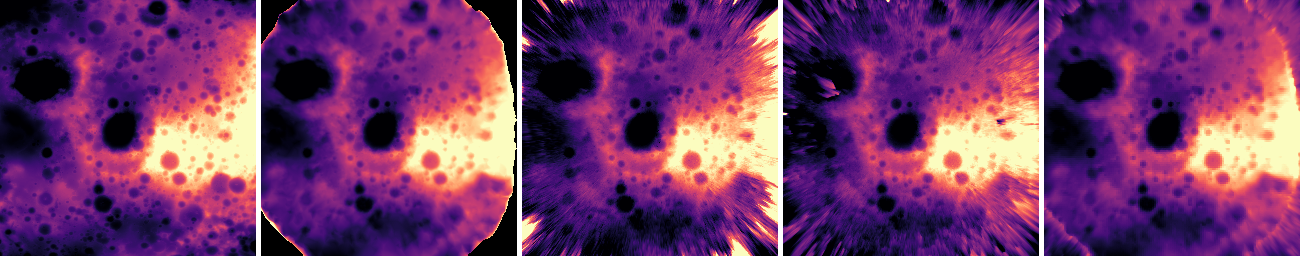}
    \vspace{-2pt}
    \\
    \makebox[\textwidth][c]{%
        \makebox[0.19\textwidth][c]{\textbf{GT}}%
        \hfill
        \makebox[0.19\textwidth][c]{\textbf{Metashape}}%
        \hfill
        \makebox[0.19\textwidth][c]{\textbf{Nerfacto}}%
        \hfill
        \makebox[0.19\textwidth][c]{\shortstack{\textbf{ours - w/o MVS} \\ \textbf{supervision}}}
        \hfill
        \makebox[0.19\textwidth][c]{\shortstack{\textbf{ours - with MVS} \\ \textbf{ supervision}}}
    }
    \vspace{-5pt}
    \caption{
    Qualitative comparison of DEM reconstructions from simulated lunar fisheye descent imagery against a high-resolution mesh ground truth. The height is calculated from the top-most camera. Our method achieves improved surface coverage while maintaining satisfactory elevation accuracy.
    }
    \label{fig:lunar}
\end{figure*}
\begin{figure*}[t]
    \centering
    \includegraphics[width=\textwidth]{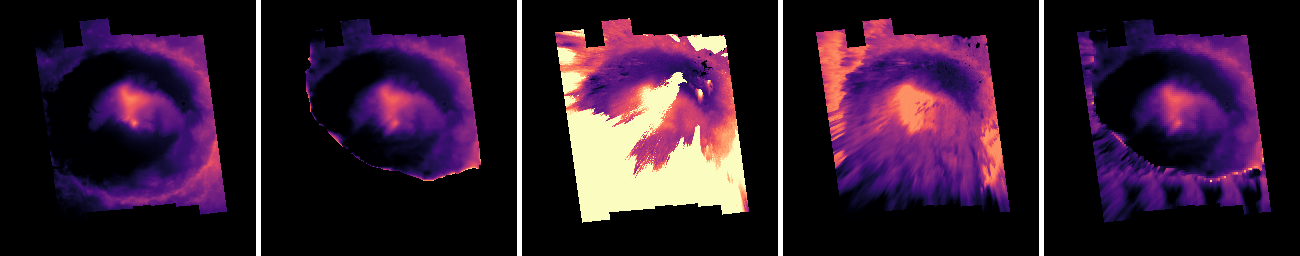}
    \vspace{-2pt}
    \\
    \makebox[\textwidth][c]{%
        \makebox[0.19\textwidth][c]{\textbf{GT}}%
        \hfill
        \makebox[0.19\textwidth][c]{\textbf{Metashape}}%
        \hfill
        \makebox[0.19\textwidth][c]{\textbf{Nerfacto}}%
        \hfill
        \makebox[0.19\textwidth][c]{\shortstack{\textbf{ours - w/o MVS} \\ \textbf{supervision}}}
        \hfill
        \makebox[0.19\textwidth][c]{\shortstack{\textbf{ours - with MVS} \\ \textbf{ supervision}}}
    }
    \vspace{-5pt}
    \caption{
    Qualitative comparison of reconstructed DEMs from simulated Mars fisheye descent imagery, using a high-resolution mesh as ground truth. Our method achieves superior surface coverage while maintaining satisfactory elevation accuracy. The height is calculated from the top-most camera. Notably, Nerfacto suffers from floating artifacts, whereas our approach effectively suppresses these floaters, resulting in a smoother and more coherent surface reconstruction.
    }
    \label{fig:mars}
\end{figure*}

Our study builds on Nerfacto and develops a novel approach specialized for the task of reconstructing DEMs from wide-angle descent images. Our objective is to learn a continuous function $\mathrm{HeightField}$ that maps planar surface coordinates $\mathbf{(x, y)}$ to their corresponding elevation values. The input data are fisheye RGB descent images, along with their camera intrinsics and extrinsics.

\paragraph{Scene representation.} 
Our approach simultaneously optimizes two representations of the planetary surface.
The first is a standard NeRF representation (Sec.~\ref{subsec:method-nerf}), i.e.~a neural network maps 3D coordinates to density.
The second is the function \(\mathrm{HeightField}\) that maps 2D coordinates in a ground plane to elevations.
These two representations have complementary properties.
NeRF is less prone to local optima during optimization, while 
the heightmap encodes the prior knowledge that planetary surfaces are typically solid and lacking significant overhangs, avoiding pathological solutions with floating geometry.
Importantly, we can also directly read off elevations from \(\mathrm{HeightField}\) to define a DEM.

In the NeRF representation (purple in Figure~\ref{fig:pipeline}), the coordinates of a given position $(\mathbf{x,y,z})$ are passed through a hash embedding then into \(\mathrm{MLP_\rho}\) to predict density \(\mathbf{\rho_{xyz;\text{A}}}\) and density embedding of the point \(\mathbf{emb_{xyz}}\). Then, density embedding is passed into \(\mathrm{MLP_{c}}\) to predict viewing direction independent colour of the point, which is then multiplied by a coefficient $\mathrm{light_d}$ given by the Hapke shading model~\cite{Sato2014Hapke} to get \(\mathbf{c_{xyz;A}}\).

In the heightmap representation (green in Figure~\ref{fig:pipeline}), assuming that the surface is roughly perpendicular to the $z$-axis, the $\mathbf{(x,y)}$ coordinates of the point are passed onto our \(\mathrm{HeightField}\) to predict the height of the surface \(\mathbf{h_{xy}}\) and the density embedding of the x,y column \(\mathbf{emb_{xy}}\). The raw height output by the MLP is scaled and bounded by applying a scaled hyperbolic tangent. Then density \(\mathbf{\rho_{xyz;B}}\) is calculated as:
\begin{equation}
\mathbf{\rho_{xyz;B}} = k_2\,\mathrm{sigmoid}\,(k_1\,(\mathbf{h_{xy}}- \mathbf{z})). 
\end{equation}
\(\mathbf{emb_{xy}}\) is passed into \(\mathrm{MLP_{c}}\) to predict the constant colour of the column; this is then multiplied by Hapke lighting $\mathrm{light_d}$ to get the colour of the point \(\mathbf{c_{xyz;B}}\). 

\paragraph{Optimization.}
Both representations define a density and colour at each point in 3D space, and we can render them to pixels following Eqs.~2--3.
Similar to NeRF, we minimize the difference between the rendered reconstruction and the original descent imagery, on minibatches of randomly sampled pixels.
We also encourage the NeRF and heightmap to express similar geometry, allowing the heightmap to regularize the NeRF and distill geometric information from it.

For each input wide-angle image $\mathbf{I_n}$, $\mathbf{n = 1,\dots,N_\text{img}}$, we cast rays $\mathbf{r}_n$ into the scene, accounting for the \textit{fisheye distortion} as shown in Figure~\ref{fig:fisheye_rays}, considering only pixels marked as valid by $\mathbf{m_n}$. The ray direction (d) is computed by mapping the pixel to the normalized camera frame using fisheye intrinsics and distortion: $\tilde{\mathbf{d}}_{n} = f_\text{fisheye}(u,v; \mathbf{K}_n, \boldsymbol{\xi}_n)$, and then rotating to world coordinates using the extrinsic rotation $\mathbf{R}_n$: $\mathbf{d}_{n} = \mathbf{R}_n \, \tilde{\mathbf{d}}_{n}$. We can then sample points $x$ along the rays, denoting each sampled point as $(\mathbf{x, y, z})$. Sampled points are spatially distorted as described in Eq.~5 to allow capturing distant surface points, and $\mathbf{z}$ is bounded by approximate $z_{min}$, $z_{max}$ values depending on the planetary curvature and topography.

We first define separate RGB loss terms for the rendering of each scene representation, as follows: 
\begin{equation}
    \begin{split}
\mathcal{L}_{c;A/B} &=
\lambda_{c,A/B}\, \mathrm{MSE}(\mathbf{c_{u,v}}, \mathbf{c_{u,v;A/B}})
\end{split}
\end{equation}

Secondly, we incorporate a distortion loss to regularize spatial inconsistencies in the reconstructed points (floaters) in the NeRF pipeline, as shown in Eq.~5~\cite{barron2022mip}. When applying this to fisheye images, the central pixels have a small angular spread, leading to small 3D sampling intervals and therefore a naturally lower distortion loss contribution. Edge pixels in a fisheye image are spatially compressed but correspond to a large angular spread, producing elongated sampling intervals along rays that can disproportionately amplify the distortion loss and introduce reconstruction errors, as shown in section~\ref{sec:angle-dist}. Replacing the standard distance-based distortion loss (Eq.~5) with an angle-based formulation mitigates this issue, leading to more stable and accurate results. Our refined distortion loss is as follows:
\begin{equation}
\mathcal{L}_{dist;A} = \lambda_{dist} \mathbf{\cos{\theta_d}}\ \iint \mathbf{w(u)w(v)|u-v|\,du\,dv} 
\end{equation}

In order to transfer geometric knowledge from the NeRF to the heightmap, we define a grid at the height ($\mathbf{h_k}$) of the highest camera, covering the required spatial region. Let $\mathbf{q}_{ij}$ denote the sampled point in the $\mathbf{(i,j)}$-th grid cell. We perform stratified sampling within each cell to obtain a ray origin $\mathbf{o}_{ij}$, and cast a vertical ray along the $z$-axis with direction $\mathbf{d} = (0,0,1)$. 

Along each ray, we sample points and compute the expected depth $\mathbf{\delta_{ij}}$ using the NeRF $\mathrm{MLP}_\mathbf{\rho}$, and the corresponding height using the $\mathrm{HeightField}$ representation:
\begin{equation}
\mathbf{\delta_{ij}} = \mathbb{E}_{z \sim \text{Ray}(\mathbf{o}_{ij}, \mathbf{d})} \big[ \mathrm{MLP}_\rho(\mathbf{o}_{ij} + z \mathbf{d}) \big] 
\end{equation}
\begin{equation}
\quad
h_{ij} = \mathrm{HeightField}(\mathbf{o}_{ij}).
\end{equation}

We define a loss term as follows:
\begin{equation}
    \begin{split}
\mathcal{L}_{height;AB} &=
\lambda_{height}\, \mathrm{L1}(\mathbf{h_k} + \mathbf{\delta_{ij}},\mathbf{h_{ij}})
\end{split}
\end{equation}
We use L1 loss here to reduce the effect of outliers.

We train the parameters of \(\mathrm{HeightField}\), \(\mathrm{MLP_d}\), and \(\mathrm{MLP_{c}}\) to minimize the loss function:
\begin{equation}
    \mathcal{L} = \mathcal{L}_{c;A} +\mathcal{L}_{c;B} + \mathcal{L}_{dist;A} + \mathcal{L}_{height;AB}
\end{equation}
In summary, $\mathcal{L}_{\mathrm{c;A}}$ and $\mathcal{L}_{\mathrm{c;B}}$ guide learning from image data, $\mathcal{L}_{\text{dist;A}}$ enforces surface smoothness, and $\mathcal{L}_{\mathrm{height;AB}}$ helps the $\mathrm{HeightField}$ capture geometric information through $\mathrm{MLP_\rho}$.

Optionally, we allow our \textit{HeightField} to be supervised by an MVS reconstruction. Specifically, we add the following auxiliary loss term to $\mathcal{L}$, with a higher weight $\lambda_{\text{MVS}}$ during the early stages of training, which is then reduced as the model learns from this supervision:
\begin{equation}
\mathcal{L}_{\text{MVS}} =
\lambda_{\text{MVS}} \, \mathrm{L1}\!\left(\mathbf{H}^{\text{MVS}}_{ij}, \mathbf{h}_{ij}\right).
\end{equation}

\section{Experiments}

\begin{table}[t]
    \centering
    \small
    \resizebox{\linewidth}{!}{ 
    \begin{tabular}{@{}l l c c c c c c@{}}
        \toprule
        Dataset & Model 
        & AED (m) $\downarrow$ 
        & RED (m) $\downarrow$ 
        & Coverage@0.1 (\%) $\uparrow$  \\
        \midrule
        \multirow{5}{*}{Moon}
        & Metashape & 196.49 & \textbf{23.05}  & 92.30\\
        & Nerfacto & 198.16 & 60.30 & 99.16 &  \\
        & Ours (w/o MVS sup.)      & 188.98 & 46.63 &  99.52  \\
        & Ours (with MVS sup.)      & \textbf{117.84} & 27.46 &  \textbf{100.00}  \\
        \midrule
        \multirow{5}{*}{Mars}
        & Metashape  & 1388.09 & \textbf{72.64} &   70.32 \\
        & Nerfacto & 9549.76 & 2022.66  & 49.88 \\
        & Ours (w/o MVS sup.)     & 1560.59 & 92.53  & 95.58 \\
        & Ours (with MVS sup.)      & \textbf{597.98} & 89.34 &  \textbf{99.93}  \\
        \bottomrule
    \end{tabular}}
    \caption{Quantitative comparison with Metashape and Nerfacto on lunar and Mars datasets. AED and RED were computed after filling NaNs with the spatially nearest available point, while Coverage@0.1 counts a point as correct only if the relative elevation difference is below 0.1, treating NaNs as incorrect. Ours has shown better Coverage@0.1 compared to baselines.}
    \label{tab:quan_results}
\end{table}

\begin{figure*}[t]
    \centering
    \includegraphics[width=\textwidth]{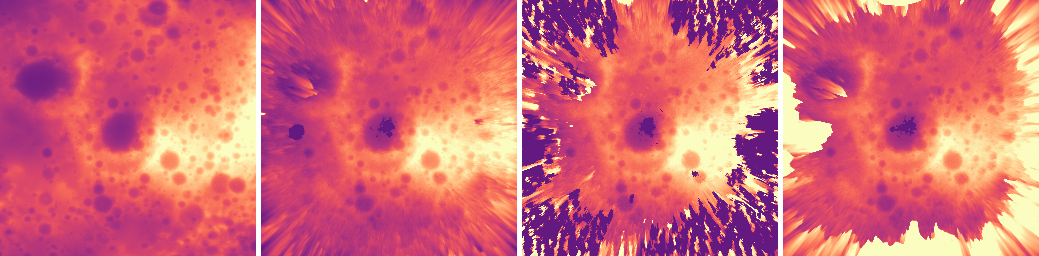}
    \vspace{-2pt}
    \\
    \makebox[\textwidth][c]{%
        \makebox[0.24\textwidth][c]{\textbf{GT}}%
        \hfill
        \makebox[0.24\textwidth][c]{\shortstack{\textbf{Ours: heightmap} \\ \textbf{ representation}}}%
        \hfill
        \makebox[0.24\textwidth][c]{\shortstack{\textbf{with MiP-NeRF 360's } \\ \textbf{distortion loss}}}%
        \hfill
        \makebox[0.24\textwidth][c]{\shortstack{\textbf{with NeRF-like} \\ \textbf{direction dependent colour}}}%
    }
    \vspace{-5pt}
    \caption{
    Qualitative comparisons of DEMs using simulated lunar fisheye descent images, with a high-resolution mesh as ground truth. Our full model demonstrates superior coverage while maintaining high-quality reconstruction compared to two ablations.
    }
    \label{fig:ablations}
\end{figure*}

We evaluate our method against modern baselines from both classical MVS (Metashape) and NeRF‑based reconstruction (Nerfacto) on two realistic datasets.

\paragraph{Datasets.}
For evaluation, we use two simulated descent sequences, one on the Moon and one on Mars. These model the descent of a $150^\circ$ fisheye camera in lunar and Mars scenes using SIMply~\cite{Brydon2025SIMply}. The lunar scene was constructed from an existing high-resolution DEM mosaic~\cite{Scholten2012GLD100} combined with resolved maps of the surface Hapke parameters~\cite{Sato2014Hapke}. In our evaluation, we use a region that covers approximately  $100\times100$km and 30 descent images of $1000\times1000$ resolution that descend towards the middle of the scene (Figure~\ref{fig:lunar_data}). The Mars scene was constructed from a Mars Context Camera (CTX) DEM mosaic and a CTX normal albedo mosaic, covering a $150\times220$km region surrounding Gale Crater~\cite{UCLMars3DOrbital}. We simulated 40 descent images of $1024\times1024$ resolution (Figure~\ref{fig:mars_data}). We generated masks to make the model ignore the black borders in all images when casting rays.

\paragraph{Metrics.}
We evaluate predicted heightmaps of size $256\times256$ pixels using three complementary metrics: \textit{Absolute Elevation Difference} (AED), \textit{Relative Elevation Difference} (RED), and \textit{Coverage@0.1}. AED measures the mean absolute vertical error between the predicted height map $\hat{H}$ and the ground-truth height map $H$. It is defined as
$\frac{1}{N} \sum_{p} \left| \hat{H}(p) - H(p) \right|$
where $p$ indexes valid pixels in the ground-truth height map and $N$ denotes the total number of evaluated pixels. This metric captures global elevation accuracy and penalizes systematic height offsets. RED captures local shape fidelity independently of global height offsets. For each pixel, the predicted height $\hat{H}(p)$ is first converted to a relative height:
$\hat{H}_{\mathrm{rel}}(p) = \hat{H}(p) - \mu_{\hat{H}}(p)$, where $\mu_{\hat{H}}(p)$ is the mean elevation within a 1\,km square window centered at pixel $p$. The same normalization is applied to the ground truth to obtain
$H_{\mathrm{rel}}(p) = H(p) - \mu_{H}(p)$.
The RED metric is then computed analogously to AED by comparing $\hat{H}_{\mathrm{rel}}$ and $H_{\mathrm{rel}}$. This formulation emphasizes the accuracy of local terrain variations while discounting global elevation shifts. When calculating these two metrics, we replaced invalid pixels with the $\hat{H}$ of the closest valid pixel. 
\textit{coverage@0.1} measures the proportion of valid points in $\hat{H}$ where the Elevation Difference relative to the ground is less than 0.1. It is defined as:
\begin{equation}
\text{Coverage@0.1} =
\frac{
\sum_{p} \mathbf{1}\Big(
\frac{|\hat{H} - H|}{|H|} \le 0.1
\Big)
}{
N
}
\end{equation}
In general, high coverage is desirable since it maximizes the benefit of capturing wide-angle images during the descent.

\subsection{Baselines}

\paragraph{Metashape.} 
A widely used practical implementation of the pipeline that combines Structure-from-Motion (SfM)~\cite{ullman1979interpretation} and Multi-View Stereo (MVS)~\cite{seitz2006comparison} is Agisoft Metashape~\cite{AgisoftMetashape}. It integrates feature detection and matching, bundle adjustment, dense depth-map estimation, point cloud densification, mesh reconstruction, and texture mapping into a unified photogrammetry framework. It automates the complete SfM–MVS workflow, enabling end-to-end 3D reconstruction from image alignment to textured mesh generation, and is extensively adopted in applications such as robotics mapping, cultural heritage documentation, surveying, and computer vision research.

\paragraph{Nerfacto.}
Nerfacto~\cite{nerfstudio2023} is a neural rendering method built on the NeRF~\cite{Mildenhall2020-hc} framework for efficiently reconstructing 3D scenes from images. It extends the standard NeRF formulation with techniques from Mip-NeRF 360~\cite{barron2022mip} to handle unbounded, large-scale scenes, and from Instant-NGP~\cite{muller2022instant} to improve computational efficiency. This combination allows Nerfacto to produce detailed geometry and high-quality novel views, making it well-suited for planetary surface reconstruction from wide-angle descent imagery. As our baseline, we further adapted the model to accept fisheye images and per-image masks as inputs.

\subsection{Comparison with Baselines}

\paragraph{Without MVS Supervision.}
As seen in Figure~\ref{fig:lunar} and Figure~\ref{fig:mars}, our approach delivers broader coverage on both lunar and Mars data with a significant precision of fine-scale reconstruction. Additionally, we can see in Figure~\ref{fig:mars} that ours has predicted a smoother surface without floaters compared to Nerfacto.  Table~\ref{tab:quan_results} further demonstrates that our method outperforms the baselines in terms of coverage, for relative elevation differences below 0.1. Notably, ours achieves superior performance compared to Nerfacto across all metrics on both datasets: AED, RED, and Coverage@0.1, demonstrating that our enhancements are effective. Ours also shows a performance increase of ~7.22\%  on lunar data with respect to Coverage@0.1, and a very substantial increase of 25.26\% on Mars data. Note that the empty regions of their reconstruction were filled with the spatially nearest value; This is disadvantageous for NeRF-based methods, including ours, as they may predict outliers rather than leaving invalid regions as NaN. However, Metashape maintains accurate reconstruction quality in areas of the scene with high image overlap.

\paragraph{With MVS Supervision.}
We optionally incorporate supervision from a classical multi-view stereo (MVS) method by using \textit{Metashape}~\cite{AgisoftMetashape} to define the auxiliary loss $\mathcal{L}_{\text{MVS}}$. As shown in Table~\ref{tab:quan_results}, incorporating the supervision of Metashape into our method leads to significantly improved performance in terms of AED and Coverage@0.1. This improvement stems from effectively combining the broad surface coverage of our approach with the high geometric accuracy of Metashape in regions with substantial image overlap, as illustrated in Figures~\ref{fig:lunar} and~\ref{fig:mars}.

\subsection{Ablations}

All ablation studies are conducted using our model \emph{without MVS supervision}.

\paragraph{Angle-based distortion loss.}\label{sec:angle-dist}
We introduce an angle-based distortion loss that adapts the regularization strength according to ray direction (Eq.~8). This formulation increases regularization for central rays, while relaxing it for edge rays, resulting in more accurate and stable surface reconstruction.
To assess its impact, we optimized an additional model using the standard distortion loss with the same weighting factor on lunar data. As shown in Figure~\ref{fig:ablations}, the model without angle-based distortion loss produces lower-quality elevation estimates near the edges. It achieves only 69.63\% Coverage@0.1, demonstrating reduced spatial completeness compared to our model.

\paragraph{Hapke lighting.}
Typically, NeRF passes direction encoding to the MLP that generates the colour of each sampled point to implicitly model direction-dependent colour. We instead use a physically-realistic shading model designed specifically to model planetary surfaces~\cite{Sato2014Hapke}, that calculates reflected colour based on incident light, viewing, and approximate surface normal directions, considering that we know the direction of sunlight at the time of the descent. To evaluate the effectiveness of this additional physical prior, we optimized another model on lunar data with implicit handling of lighting by passing direction encoding onto the model. We can see in Figure~\ref{fig:ablations} that a more accurate lighting model is important for high-quality DEMs. Quantitatively, this model shows only 88.55\% Coverage@0.1 compared to 99.52\% coverage of our full model.

\section{Conclusion}

We have presented the first study of modern neural 3D reconstruction techniques for building digital elevation models using wide-angle descent imagery.
As well as evaluating existing methods, we proposed a novel technique specialized for this domain, incorporating a heightmap representation and physical priors.
Quantitative and qualitative evaluations of Metashape, Nerfacto, and our proposed approach on lunar and Mars datasets show that these methods have promise for this domain. Our method achieves greater coverage while maintaining high accuracy in DEM reconstruction. 
{
    \small
    \bibliographystyle{ieeenat_fullname}
    \bibliography{main}
}


\end{document}